%% file: arxiv.tex
\documentclass[letterpaper, 10 pt, conference]{ieeeconf}  

\IEEEoverridecommandlockouts                              

\usepackage{amsmath}
\usepackage{amssymb}
\usepackage{float}
\usepackage{graphicx}
\usepackage{xcolor}

\definecolor{draftblue}{RGB}{73,103,128}

\definecolor{comparegreen}{RGB}{35,139,69}
\definecolor{comparered}{RGB}{205,55,55}
\newcommand{\cmark}{\textcolor{comparegreen}{\ensuremath{\checkmark}}}
\newcommand{\xmark}{\textcolor{comparered}{\ensuremath{\times}}}
\usepackage{hyperref}

\title{\LARGE \bf
EgoAlign: Bridging the Human-Humanoid Gap for Long-Range Loco-Manipulation
}

\input{authors}

\begin{document}

\maketitle
\thispagestyle{empty}
\pagestyle{empty}

\begin{abstract}
Egocentric human demonstrations offer an accessible source of task experience, but differences in body scale and controller response, together with missing robot states, limit their value as humanoid training supervision. We present EgoAlign, a data-construction framework that converts these demonstrations into action and state supervision compatible with a general-purpose, continuous whole-body controller, without collecting physical-robot demonstrations. Using the target-robot model and simulator, EgoAlign guides demonstration collection through execution feedback. It preserves locomotion references for visually guided periodic stepping while adapting upper-body interaction geometry through scale alignment and controller-in-the-loop refinement. A final causal replay reconstructs the corresponding robot states and motion-token labels for training with the human observations. We assess the resulting supervision by fine-tuning a vision--language--action model solely on adapted human demonstrations and deploying it zero-shot on a physical humanoid. The resulting policies perform long-range object relocation, navigation to unseen goal positions, and independently evaluated foot interaction. Refinement improves simulated hand alignment and physical pickup success over kinematic alignment alone, while human collection reduces on-site acquisition time relative to teleoperation. \url{https://lambdahumanoid.github.io/EgoAlign/}
\end{abstract}

\section{INTRODUCTION}

Scaling vision-language-action (VLA) learning requires abundant, diverse, effective training data~\cite{kim2024openvla,physicalintelligence2025pi05}. Physical-robot teleoperation is costly and exposes robots to interaction failures. Egocentric human demonstrations provide navigation and interaction experience without operating a physical robot~\cite{chi2024umi,shi2026egohumanoid} and could complement robot demonstrations in large-scale mixed-data training~\cite{wang2026ego2robot}. Their value, however, depends on target-embodiment supervision quality: a successful human action does not directly provide a suitable robot training example.

Here, quality means human-derived actions and corresponding robot states suitable for policy learning. Body-scale and controller-response differences can make realized hand trajectories deviate from human references, while recordings lack the associated robot state history. These gaps matter for long-range loco-manipulation, which combines distant navigation, sustained interaction during locomotion, and local contact~\cite{wei2026psi0}. How can we bridge these gaps to make egocentric human demonstrations effective supervision for long-range humanoid loco-manipulation?

We present EgoAlign, which improves the target-robot compatibility of human demonstrations through motion adaptation and causal state reconstruction. It constructs VLA supervision for SONIC's public, general-purpose, continuous whole-body interface~\cite{luo2025sonic}, so learned policies can use leg motions for both locomotion and foot interaction without task-specific controller training (Table~\ref{tab:closest_systems}). During locomotion, visual feedback guides periodic stepping toward goals without requiring frame-by-frame matching of human and robot displacement or speed. Hand--object interaction requires local geometric alignment, so we adapt upper-body motion while preserving global travel and lower-body references. Yet kinematic alignment alone cannot ensure controller-realized accuracy. During collection, SONIC--MuJoCo feedback~\cite{todorov2012mujoco} guides correction of suppressed motions. Offline, scale alignment combines kinematic alignment with controller-in-the-loop refinement to correct residual errors in realized upper-body motion. Because motion corrections change the resulting robot states, final causal replay reconstructs corresponding states and motion-token labels for training with the original human observations. This uses the target robot's model, controller, and simulation, but no physical-robot demonstrations.

\begin{table}[t]
\caption{Robot-free humanoid learning systems.
EE. Capture: end-effector collection. Navigation: reported held-out settings.
\cmark/\xmark: demonstrated/not demonstrated.}
\label{tab:closest_systems}
\centering
\footnotesize
\renewcommand{\arraystretch}{1.2}
\setlength{\tabcolsep}{2pt}
\begin{tabular*}{\columnwidth}{@{\extracolsep{\fill}}lcccc@{}}
\hline
\rule{0pt}{3.5ex}Method & \shortstack{EE.\\Capture} & \shortstack{Pub. \& Conti.\\General WBC} & \shortstack{Nav.\\Generalization} & \shortstack{Foot\\Interact.} \\
\hline
HuMI~\cite{nai2026humi} & UMI & \xmark & Not reported & \xmark \\
BifrostUMI~\cite{wang2026bifrostumi} & UMI & \xmark & Not reported & \xmark \\
EgoHumanoid~\cite{shi2026egohumanoid} & Hands & \xmark & Cross-scene & \xmark \\
\hline
\textbf{EgoAlign (ours)} & \textbf{Hands} & \cmark & \textbf{Goal-position} & \cmark \\
\hline
\end{tabular*}
\vspace{-4pt}
\end{table}

We use human-only task training followed by zero-shot physical deployment as a functional test of the adapted data's training value. Using the $\pi_{0.5}$ architecture~\cite{physicalintelligence2025pi05} with a pretrained vision--language backbone, we train on these demonstrations and deploy in closed loop without task-specific physical-robot demonstrations or robot-side adaptation. Evaluation covers roughly $10$-m end-to-end object relocation and independently tested position-generalized navigation and foot interaction. Our alignment reduces simulated right-palm error from $14.27$ to $1.65$ cm and increases physical object-pickup success from 0\% to 90\%. Same-stack ablations assess navigation data, level-view perception, and state reconstruction, while a teleoperation comparison measures on-site collection efficiency.

Our contributions are threefold:
\begin{itemize}
\item We introduce EgoAlign, a robot-free framework that constructs robot-compatible action and state supervision from egocentric human demonstrations for VLA learning through a general-purpose, continuous whole-body control interface.
\item We develop motion adaptation through dynamics-guided collection and upper-body scale alignment, followed by causal state reconstruction to pair adapted actions with their corresponding robot states.
\item We validate the supervision through human-only training and zero-shot physical deployment on long-range object relocation, position-generalized navigation, and independently evaluated foot interaction, complemented by ablations, alignment diagnostics, and collection-efficiency measurements.
\end{itemize}

\begin{figure*}[t]
    \centering
    \includegraphics[width=\textwidth]{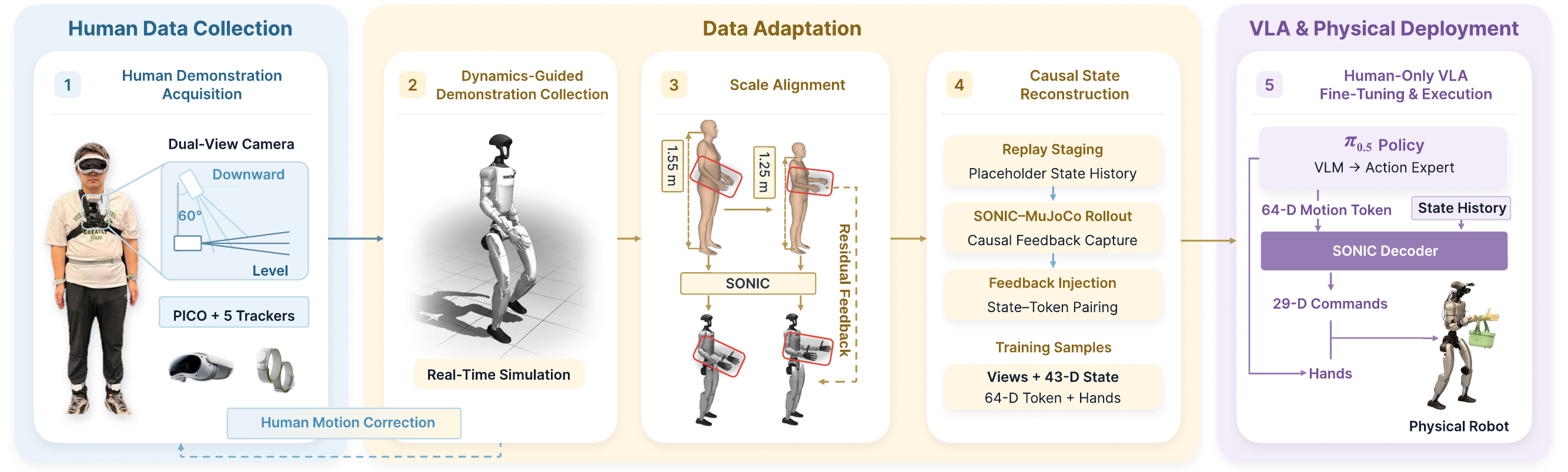}
    \caption{EgoAlign pipeline. (1) A wearable setup captures synchronized dual-view images and human motion. (2) Real-time SONIC–MuJoCo feedback guides demonstrators to adjust their motions.
    (3) Scale alignment adapts upper-body motion to human-reference hand targets through optimization and replay.
    (4) State reconstruction pairs pre-action robot states and
    same-tick motion tokens.
    (5) Human-only task training produces a $\pi_{0.5}$ policy for zero-shot robot deployment, predicting 64-D motion tokens and hand commands from live observations; SONIC decodes tokens with state history into body commands.
    Dashed arrows indicate collection-time and offline feedback.}
    \label{fig:pipeline}
    \vspace{-4pt}
\end{figure*}

\section{RELATED WORK}

\subsection{Vision-Language-Action Models}

Open vision-language-action (VLA) models support robot pretraining and task adaptation. OpenVLA~\cite{kim2024openvla} trains a generalist policy on heterogeneous robot demonstrations. StarVLA~\cite{starvla2026} provides a modular framework for comparing VLA backbones, action heads, and training recipes. The $\pi$ family progresses from the flow-matching $\pi_0$~\cite{black2024pi0}, through the open-world mobile-manipulation $\pi_{0.5}$~\cite{physicalintelligence2025pi05}, to the steerable generalist $\pi_{0.7}$~\cite{physicalintelligence2026pi07}. JoyAI-RA~0.1~\cite{zhang2026joyaira} instead integrates web, simulation, egocentric human, and robot data with an explicitly unified action space. Recent work further broadens the observation space to sound-centric manipulation~\cite{nie2026towards}. Most relevant to humanoids, $\Psi_0$~\cite{wei2026psi0} pretrains on egocentric human video and post-trains a whole-body action expert on humanoid trajectories for loco-manipulation. Complementary work studies spatial representations for robotic perception~\cite{deng2026prosgnerf,deng2025best3dscenerepresentation}, whereas we focus on constructing robot-compatible supervision from human demonstrations.

\subsection{Robot-Free Human Demonstrations}

Robot-free interfaces such as UMI~\cite{chi2024umi} and FastUMI~\cite{zhaxizhuoma2025fastumi} reduce the hardware cost of collecting executable manipulation demonstrations. Complementary efforts scale natural egocentric experience: EgoLive~\cite{li2026egolive} emphasizes annotated real-world routines, EgoVerse~\cite{punamiya2026egoverse} standardizes diverse data and cross-laboratory evaluation, and EgoScale~\cite{zheng2026egoscale} studies predictable scaling of human data for dexterous transfer. EgoMimic~\cite{kareer2024egomimic} jointly aligns human and robot demonstrations, whereas HumanEgo~\cite{wang2026humanego} uses interaction-centric representations to transfer from minutes of human video without robot training data. These results motivate human demonstrations as scalable supervision while highlighting the remaining visual and kinematic embodiment gaps.

\subsection{Humanoid Loco-Manipulation and Whole-Body Control}

OmniH2O~\cite{he2024omnih2o} and HOMIE~\cite{ben2025homie} provide whole-body teleoperation and control. HuMI~\cite{nai2026humi} and BifrostUMI~\cite{wang2026bifrostumi} learn from handheld UMI demonstrations using task-space policies and learned motion trackers. HuMI's controller is not general-purpose, while BifrostUMI's tracker implementation is unreleased at the time of writing. EgoHumanoid~\cite{shi2026egohumanoid} aligns egocentric views and evaluates both human-only training and robot-data co-training, but exposes discrete lower-body primitives to its VLA. TANGO~\cite{li2026tango} constructs whole-body navigation supervision from planned and edited motions verified by SONIC, while VLK~\cite{wang2026vlk} synthesizes whole-body interactions and paired egocentric views in reconstructed scenes. EgoAlign instead constructs controller-compatible action and state supervision from recorded human demonstrations through motion adaptation and causal state reconstruction. Using SONIC~\cite{luo2025sonic} as a shared continuous whole-body interface, the learned policies support navigation and foot interaction. Table~\ref{tab:closest_systems} summarizes the collection, control, and demonstrated capability differences.

\section{METHOD}

\subsection{Problem Formulation and System Overview}

For a language instruction $\ell$, human demonstrations $\mathcal{D}_{H}$ contain synchronized downward- and level-view images $(\mathbf{I}^{\mathrm{down}},\mathbf{I}^{\mathrm{level}})$, body motion, and hand commands. We construct robot-compatible state--action supervision for fine-tuning a VLA policy $\pi_\theta(\mathbf{a}_{t:t+49}\mid\mathbf{I}^{\mathrm{down}}_t,\mathbf{I}^{\mathrm{level}}_t, \mathbf{s}^{G}_t,\ell)$ solely on $\mathcal{D}_{H}$, where $\mathbf{s}^{G}_t$ is the current robot state, $\mathbf{a}_t$ one step of the predicted 50-step action chunk, and $\theta$ the trainable parameters. Labels $H$ and $G$ denote human and robot quantities; $t$ indexes time.

EgoAlign requires a target-robot model and a continuous WBC supporting motion encoding, state-history-conditioned decoding, and simulated replay. In our SONIC implementation~\cite{luo2025sonic}, the human reference $\mathbf{r}^{H}_{t}$ comprises SMPL joints, root orientation, and tracked wrist targets. The encoder maps a reference window to a motion token,
\begin{equation}
\mathbf{m}^{S}_{t}=
E_{S}(\mathbf{r}^{H}_{t-K+1:t},\mathbf{R}^{G}_t)\in\mathbb{R}^{64},
\label{eq:sonic_encoder}
\end{equation}
where $K=10$ and $\mathbf{R}^{G}_t$ is the current robot-base orientation used to anchor the reference during rollout.

The VLA predicts these tokens rather than joint targets. SONIC decodes each token using a causal state history
\begin{equation}
\mathcal{H}^{G}_{t}=\{\boldsymbol{\omega}^{G},\mathbf{q}^{G},
\dot{\mathbf{q}}^{G},\mathbf{g}^{G}\}_{\leq t}
\cup\{\mathbf{u}^{G}\}_{<t},
\end{equation}
which contains measured base angular velocity, body joint position and velocity, projected gravity, and preceding actions over the controller's state-history window. The resulting normalized whole-body action is
\begin{equation}
\mathbf{u}^{G}_{t}=D_{S}(\mathbf{m}^{S}_{t},\mathcal{H}^{G}_{t})
\in\mathbb{R}^{29}.
\label{eq:sonic_decoder}
\end{equation}
SONIC decoder converts this action to joint-position commands; hands are controlled separately. State reconstruction follows motion adaptation (Fig.~\ref{fig:pipeline}); we introduce it first to establish the training interface.

\subsection{Causal State Reconstruction}

Equations~\eqref{eq:sonic_encoder}--\eqref{eq:sonic_decoder} imply that a human episode provides the motion reference but not the state history required to decode it. We reconstruct robot proprioceptive states and paired motion tokens under the replay dynamics for VLA training, processing each finalized motion through three stages.

In \emph{Replay Staging}, the synchronized human episode is converted into a deterministic replay reference containing body motion and hand commands, with zeroed state history as an intermediate placeholder. 

In \emph{SONIC--MuJoCo Rollout}, SONIC executes the episode on the simulated robot at $50$ Hz. At control tick $t$, the controller first reads the simulated state and updates $\mathcal{H}^{G}_{t}$, and only then evaluates Eq.~\eqref{eq:sonic_decoder}. Consequently,
\begin{equation}
\begin{aligned}
\mathbf{x}^{G}_t &= \Phi_t(\mathbf{x}^{G}_0,\mathbf{a}_{0:t-1}),\\
\mathbf{s}^{G}_t &= \mathcal{O}(\mathbf{x}^{G}_t),
\end{aligned}
\label{eq:causal_state}
\end{equation}
where $\Phi_t$ is the deterministic SONIC--MuJoCo rollout of the full simulation state $\mathbf{x}^{G}$ driven by past tokens and hand commands, and $\mathcal{O}$ extracts the $43$-D robot proprioceptive state $\mathbf{s}^{G}_t$ from $\mathbf{x}^{G}_t$ for input to the VLA. The pre-action state and the controller’s state history are constructed before executing the current action $\mathbf{a}_t$.

In \emph{Feedback Injection}, we validate row correspondence and inject the measured pre-action feedback into the training episode. The $43$-D state contains $29$ body joint positions and two $7$-D hand states. The learned VLA action is
\begin{equation}
\mathbf{a}_t=
[\,\mathbf{m}^{S}_t,\mathbf{h}^{L}_t,\mathbf{h}^{R}_t\,]
\in\mathbb{R}^{76},
\end{equation}
where $\mathbf{m}^{S}_t\in\mathbb{R}^{64}$ is captured from the SONIC encoder at the same control tick, and $\mathbf{h}^{L}_t,\mathbf{h}^{R}_t\in\mathbb{R}^{6}$ are the left and right hand-command vectors.

\subsection{Human Demonstration Acquisition}

Our motion-capture pipeline follows EgoHumanoid~\cite{shi2026egohumanoid}. A PICO 4 Ultra headset and five trackers on the wrists, ankles, and waist capture human motion. We use SONIC's PICO-to-SMPL conversion pipeline to obtain motion references in the 24-joint SMPL representation~\cite{luo2025sonic,loper2015smpl}, retaining wrist orientations and hand commands. PICO's vision-based hand recognition provides discrete grasp/non-grasp commands used in both collection and physical execution.

\begin{figure}[t]
    \centering
    \includegraphics[width=\columnwidth]{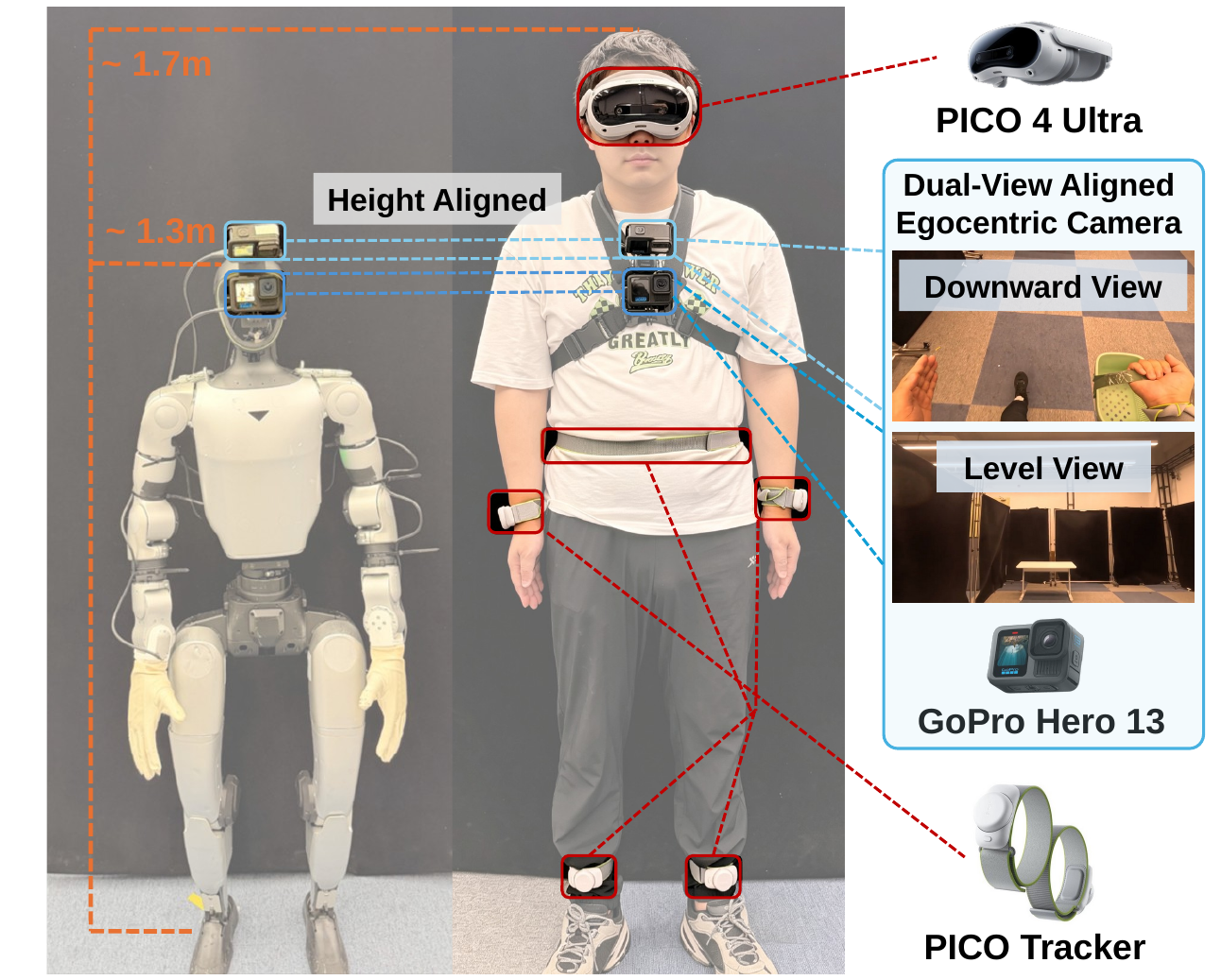}
    \caption{Human collection and robot hardware with matched camera heights.
    The downward view captures near-field interaction,
    while the level view provides distant navigation cues.}
    \label{fig:dual_view_setup}
    \vspace{-4pt}
\end{figure}

Long-range loco-manipulation requires both near-field interaction and distant navigation context. Two GoPro Hero 13 cameras provide a chest-mounted downward view $\mathbf{I}^{\mathrm{down}}$ of the hands, objects, and nearby workspace, and a horizontal level view $\mathbf{I}^{\mathrm{level}}$ of distant targets. Corresponding human and robot cameras have matched ground-relative heights (Fig.~\ref{fig:dual_view_setup}). Both streams are synchronized to the PICO trajectory before data construction and used directly for training.

\subsection{Human-to-Humanoid Motion Adaptation}

\subsubsection{Dynamics-Guided Demonstration Collection}

Accurate human tracking does not guarantee that the converted motion is expressed by the target humanoid. Scale differences, SONIC's tracking response, and robot dynamics may suppress low-amplitude actions; for example, a small human step or slight leg lift can yield almost no visible robot motion. To expose this gap during collection, we stream the recovered SMPL motion through SONIC and replay the resulting robot behavior in MuJoCo in real time. This visual feedback guides the demonstrator to repeat or amplify suppressed motions until the intended robot movement is visible in simulation, without operating a physical robot. Unlike collection-time feedback, offline alignment corrects recorded motion before state reconstruction.

\subsubsection{Scale Alignment}

\begin{figure}[t]
    \centering
    \includegraphics[width=\columnwidth]{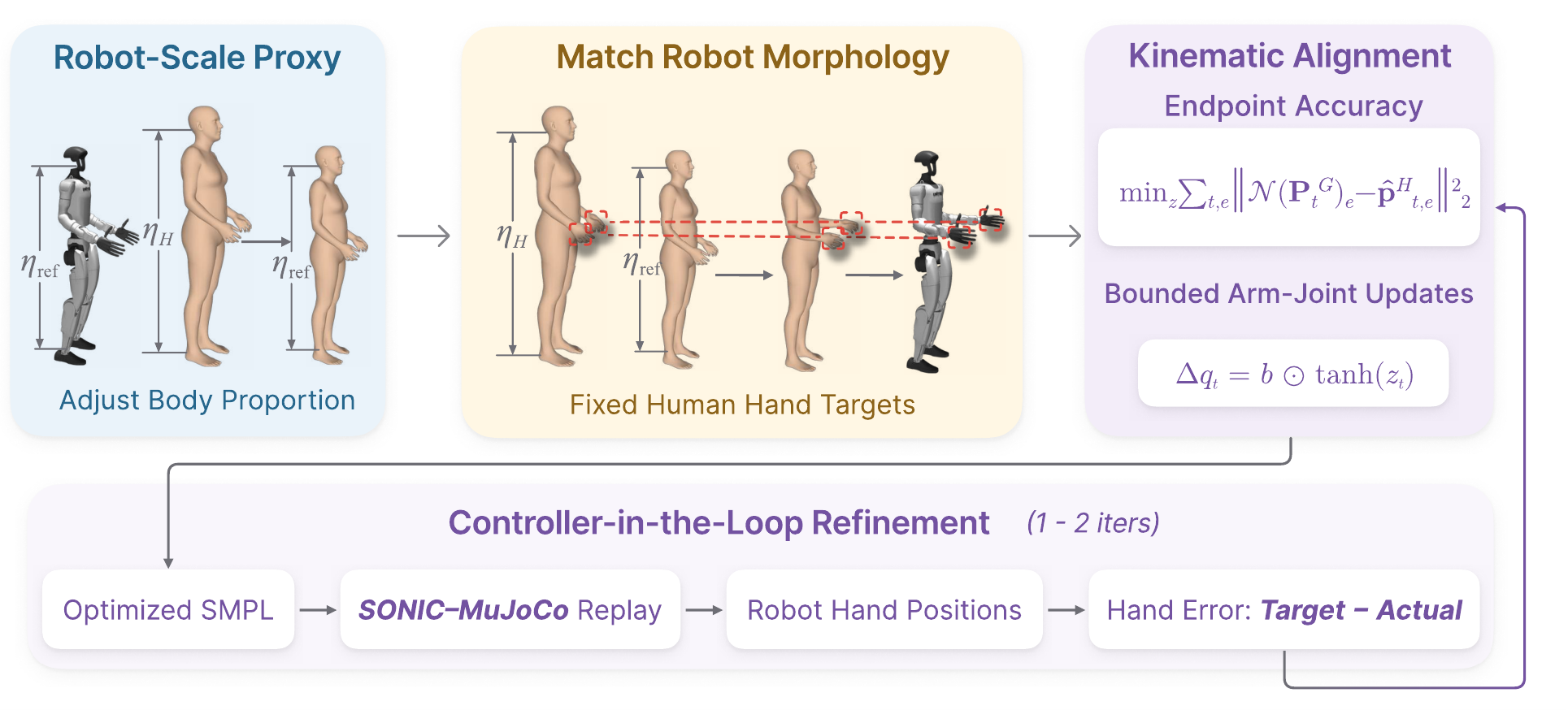}
    \par\vspace{-3pt}
    \caption{Scale alignment. Constructing a robot-scale SMPL proxy (left), matching fixed human hand targets (middle), and optimizing bounded arm-joint updates (right). The bottom loop uses SONIC--MuJoCo replay residuals for iterative refinement.}
    \label{fig:scale_alignment}
    \vspace{-8pt}
\end{figure}

Scale alignment adapts upper-body interaction geometry while preserving global root travel and lower-body motion in the SMPL reference, through kinematic alignment followed by refinement (Fig.~\ref{fig:scale_alignment}).

\emph{Kinematic Alignment.} We distinguish the human-scale reference skeleton $\mathcal{M}_H$ from an optimization-only robot-scale proxy $\mathcal{M}_G$. The human reference has canonical height $\eta_H$, measured from the lower foot joint to the head joint rather than crown to sole. Its recorded pose $\mathbf{q}_t$ defines the desired hand geometry through
\begin{equation}
\mathbf{P}_t^{H}=\operatorname{FK}(\mathbf{q}_t;\mathcal{M}_H).
\label{eq:human_reference}
\end{equation}
For a posed skeleton $\mathbf{P}_t$, we define interaction-frame coordinates as
\begin{equation}
\mathcal{N}(\mathbf{P}_{t})_j=\mathbf{P}_{t,j}-\mathbf{o}(\mathbf{P}_{t}),
\label{eq:interaction_frame}
\end{equation}
where $\mathbf{o}(\mathbf{P}_{t})$ uses the pelvis's horizontal coordinates and the lower of the two foot-joint heights as its vertical coordinate, computed separately for each input skeleton. The human-reference endpoint target is $\hat{\mathbf{p}}_{t,e}^{H}=\mathcal{N}(\mathbf{P}_{t}^{H})_e$, where $e$ indexes a selected hand endpoint. Human-reference hand targets remain fixed in the shared interaction frame. Alignment adjusts the reference pose to bring the robot's realized hand positions closer to these targets, while leaving the human images and depicted objects unchanged.

Separately, $\mathcal{M}_{G}$ retains SMPL topology but fits torso and limb lengths to the target robot's morphology (reference height $\eta_{\mathrm{ref}}$). Changing this proxy's proportions alone does not change the recorded pose or SONIC's motion input. On this proxy, we optimize a bounded pose residual $\Delta\mathbf{q}_t=\mathbf{b}\odot\tanh(\mathbf{z}_t)$, where $\mathbf{b}$ contains joint-wise angular bounds and $\mathbf{z}_t$ is the unconstrained optimization variable. With $\mathbf{P}_t^{G}=\operatorname{FK}(\mathbf{q}_t+ \Delta\mathbf{q}_t;\mathcal{M}_{G})$, the endpoint loss is
\begin{equation}
\mathcal{L}_{x}=\frac{1}{3T|\mathcal{E}|}
\sum_{t=1}^{T}\sum_{e\in\mathcal{E}}
\left\|
\mathcal{N}(\mathbf{P}_{t}^{G})_e
-\hat{\mathbf{p}}^{H}_{t,e}
\right\|_2^2.
\label{eq:endpoint_loss}
\end{equation}
Here, $T$ is the episode length, $\mathcal{E}$ is the task-selected hand-endpoint set. We regularize the residual magnitude and temporal smoothness:
\begin{equation}
\mathcal{L}=\lambda_x\mathcal{L}_{x}
+\lambda_{\mathrm{pose}}\mathcal{L}_{\mathrm{pose}}
+\lambda_{\mathrm{vel}}\mathcal{L}_{\mathrm{vel}}
+\lambda_{\mathrm{acc}}\mathcal{L}_{\mathrm{acc}},
\label{eq:scale_alignment}
\end{equation}
where $\mathcal{L}_{\mathrm{pose}}$, $\mathcal{L}_{\mathrm{vel}}$, and $\mathcal{L}_{\mathrm{acc}}$ are the mean-squared zeroth-, first-, and second-order temporal differences of $\Delta\mathbf{q}_t$, respectively. We solve Eq.~\eqref{eq:scale_alignment} with Adam. Due to space limits, detailed loss weights, joint bounds, optimization steps, and gain-update rules will be provided in the open-source code and executed configurations. Each selected hand endpoint is represented by the mean of its four non-thumb MCP roots. Optimization is restricted to the task-selected collar, shoulder, and elbow joints; all other joints, including the spine and wrist-local rotations, retain their per-frame baseline trajectories, so the spine follows its original motion rather than remaining fixed over time.

\emph{Controller-in-the-Loop Refinement.} From the kinematically aligned $\mathbf{q}^{(0)}$, SONIC--MuJoCo replay $\mathcal{R}_{S}$ yields round-$k$ endpoints and residuals in the target interaction frame:
\begin{equation}
\mathbf{y}^{(k)}=\mathcal{R}_{S}(\mathbf{q}^{(k)}), \qquad
\mathbf{e}^{(k)}_{t,e}=\hat{\mathbf{p}}^{H}_{t,e}-\mathbf{y}^{(k)}_{t,e}.
\label{eq:replay_error}
\end{equation}
For evaluation and feedback, we compensate a per-episode, per-round constant lag estimated from endpoint velocities within $\pm20$ frames, excluding unmatched boundaries, and smooth only the feedback residual $\tilde{\mathbf{e}}^{(k)}_{t,e}$ with a 21-frame, third-order Savitzky--Golay filter, without shifting images, training labels, or source indices. For assessing geometric alignment, lag compensation provides a more appropriate comparison by separating spatial mismatch from positional differences caused by controller response delay. Starting with $\mathbf{c}^{(0)}_{t,e}=\mathbf{0}$, we accumulate a bounded target correction:
\begin{equation}
\begin{aligned}
\mathbf{d}^{(k)}_{t,e} &= \operatorname{clip}_{\rho_s}
\bigl(\beta\tilde{\mathbf{e}}^{(k)}_{t,e}\oslash\boldsymbol{\kappa}^{(k)}_e\bigr),\\
\mathbf{c}^{(k+1)}_{t,e} &= \operatorname{clip}_{\rho_c}
\bigl(\mathbf{c}^{(k)}_{t,e}+\mathbf{d}^{(k)}_{t,e}\bigr).
\end{aligned}
\label{eq:feedback_update}
\end{equation}
Here, clipping bounds each correction vector's norm, $\beta=0.65$, $\rho_s=0.08$ m, and $\rho_c=0.12$ m. The per-axis response gain $\boldsymbol{\kappa}^{(k)}_e$ is shared across time for endpoint $e$, starts at one, and can be updated from successive replays using a bounded diagonal secant estimate; $\oslash$ denotes element-wise division. We then re-solve Eq.~\eqref{eq:scale_alignment} using compensated optimization targets $\hat{\mathbf{p}}^{H}_{t,e}+\mathbf{c}^{(k+1)}_{t,e}$ and replay the resulting $\mathbf{q}^{(k+1)}$. This input compensation leaves the desired human-reference targets $\hat{\mathbf{p}}^{H}_{t,e}$ unchanged. All rounds use the same baseline motion and joint mask. Starting from Round~0, we perform at most two refinements, yielding Rounds~1 and~2; trajectories that still fail the closed-loop endpoint and rollout acceptance criteria after Round~2 are excluded from training, and accepted trajectories undergo separate causal-data integrity validation. Detailed thresholds and filtering rules are omitted for space and will accompany the code release.

Finally, we regenerate the motion references from the optimized SMPL poses for causal state reconstruction. Given the endpoint and joint-mask configuration, offline optimization and replay run automatically with shared update rules and hyperparameters. Gradients pass only through the optimization skeleton, not SONIC or MuJoCo.

\subsection{Human-Only VLA Fine-Tuning and Humanoid Execution}

We fine-tune the VLA on original human images, instructions, and pre-action states from final causal replay, with paired SONIC tokens and human hand commands as action targets. During deployment, the VLA predicts actions from live robot images and the current robot state. SONIC decodes the predicted tokens using live robot state history via Eq.~\eqref{eq:sonic_decoder}, while hand commands execute separately.

\section{EXPERIMENTS}

We evaluate the training value of the constructed supervision and the efficiency of human collection through four questions: (1) How well do policies fine-tuned solely on human task demonstrations perform on a physical humanoid? (2) How do navigation data, level-view perception, state reconstruction, and scale alignment affect performance? (3) How does refinement affect simulated endpoint error and physical pickup success? (4) How much does human collection reduce the on-site acquisition time relative to teleoperated collection?

\subsection{Experimental Setup}

\subsubsection{Tasks and Data}

We use Unitree G1 (reference height $\eta_{\mathrm{ref}}=1.25$ m) with SONIC control and MuJoCo replay. Hand targets use the human-scale canonical SMPL ($\eta_H=1.55$ m), while the optimization proxy fits G1's body proportions. With cameras at approximately $1.3$ m, the downward view covers only the next $2$ m, motivating the level view.

\emph{Object relocation} comprises basket pickup, approximately $10$ m of transport, and table placement. \emph{Navigation and foot-based interaction} combines approximately $5$ m of navigation and pedal-bin opening under one jointly trained policy. The demo shows the full sequence; quantitative tests separate its subtasks, with foot interaction starting from standing.

\begin{figure*}[t]
    \centering
    \includegraphics[width=0.95\textwidth]{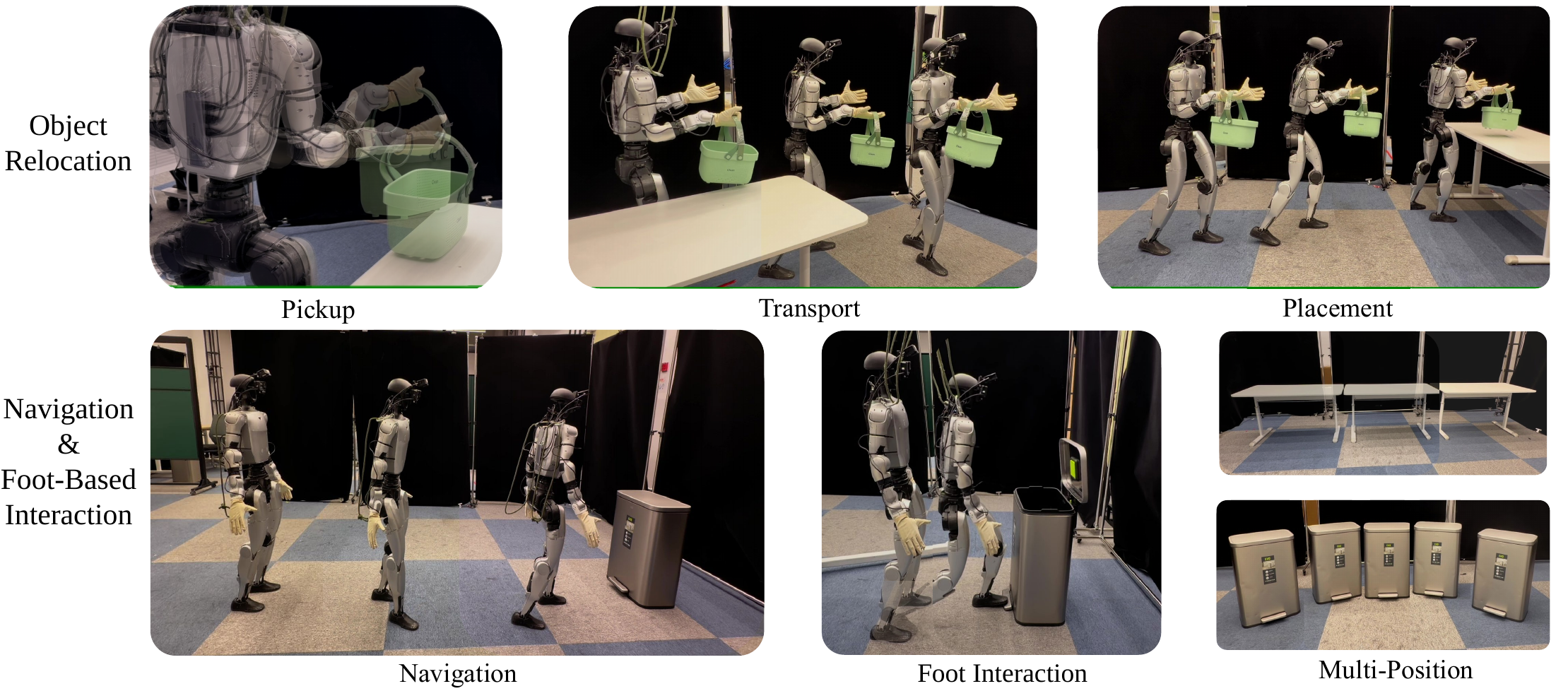}
    \caption{Physical-G1 tasks and endpoint protocols for object relocation
    and for navigation and foot-based interaction. Object relocation comprises pickup, transport,
    and placement. Navigation and foot interaction form a complete
    behavior but are evaluated independently to separate navigation and
    foot-operation performance.}
    \label{fig:task_scenes}
    \vspace{-4pt}
\end{figure*}

Object relocation and navigation use \emph{Direct} (a training-distribution endpoint directly ahead), \emph{Multi-seen} (training positions P1--P5), and \emph{Multi-unseen} (U1--U5, offset 0.5 m along either ground-plane axis from corresponding seen positions). Methods share these positions; foot interaction uses only the multi-position protocols (Fig.~\ref{fig:task_scenes}). Three demonstrators collected 200 full-trajectory object relocation demonstrations and 200 demonstrations each for navigation and foot interaction. Object relocation additionally uses 100 basket-carrying navigation demonstrations, separate from pedal-bin navigation. These counts are before filtering; approximately 90\% are retained.

Object relocation alignment optimizes the right collar, shoulder, and elbow toward the right-palm target throughout each demonstration, including transport with sustained hand--handle contact. Left-arm swing, synthesized from smoothed right-thigh sagittal motion with $20^\circ$ outward abduction, is preserved during alignment along with SMPL spine and wrist-local trajectories. Full-method physical policies and simulation share this configuration. Replay models ground contact but no task objects or hand/foot--object contacts. During object relocation replay, a grasp signal applies a downward hand-node force equivalent to the weight of a 200 g empty basket.

\subsubsection{Training and Evaluation}

Policies use the $\pi_{0.5}$ architecture, initializing only the backbone from PaliGemma-3B-PT-224~\cite{beyer2024paligemma}. Both backbone and action expert train on all available human task data for 50{,}000 steps on eight A800 GPUs (global batch 256). Variants share the sampling-weight scheme and SONIC configuration. Final checkpoints are evaluated without physical-feedback-based selection or tuning. Each 50-step chunk executes fully at 50 Hz before reobserving and requesting inference. During inference, SONIC holds the last latent motion token and continues closed-loop decoding with live states.

Each setting has 20 trials (five positions with four trials each for multi-position protocols). Trials end at completion, a two-minute timeout, or clear failure (e.g., basket drop, fall, or workspace-boundary collision); failures, interventions, and timeouts count as unsuccessful.

\subsubsection{Scoring Stages and Success Criteria}

Object relocation uses S1 Approach, S2 Lift, S3 Carry, and S4 Place. Navigation uses S1 Reach Target Region and S2 Final Alignment at a pedal-reachable pose. Foot interaction uses S1 Foot Approach, S2 Effective Pedal Press (lid opens), and S3 Stable Completion. Stages are judged manually; stages following the first failure are scored zero. All stage rates use all 20 trials. Full success is the final-stage rate, requiring every stage to pass; stage score is 100 times the mean fraction of stages completed per trial.

\subsection{Zero-Shot Loco-Manipulation on the G1}

\begin{figure}[!t]
    \centering
    \includegraphics[width=\columnwidth]{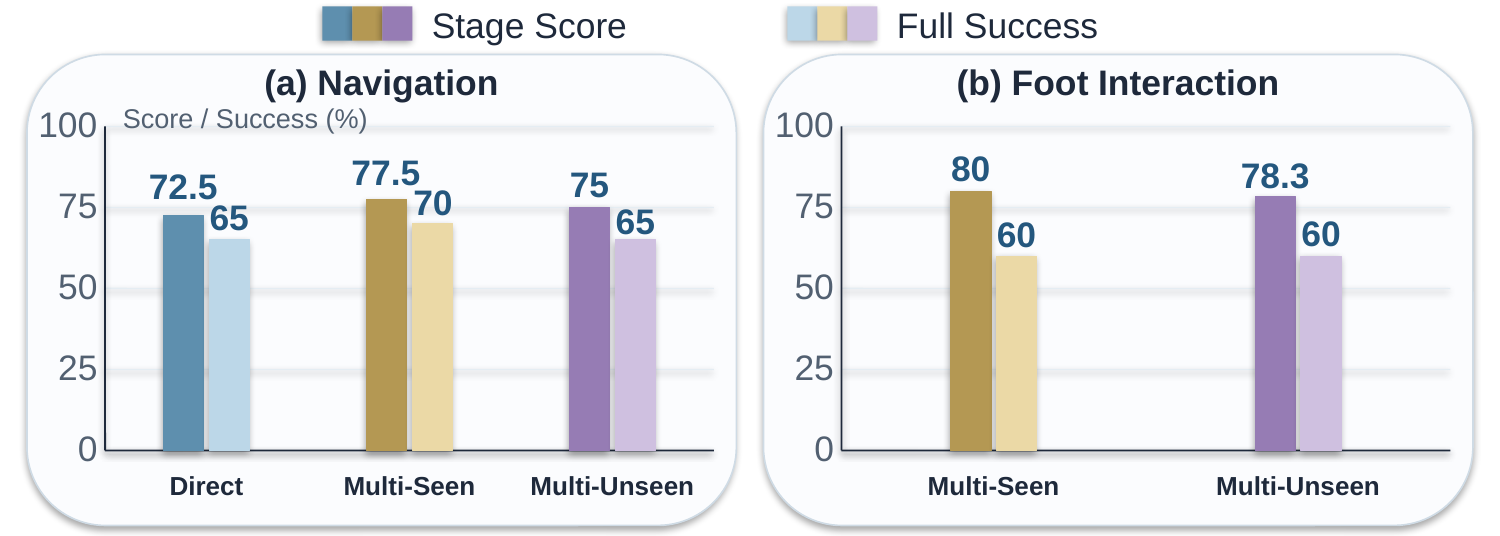}
    \caption{Stage score and full success for independently tested
    (a) navigation and (b) foot interaction, with 20 trials per setting.}
    \label{fig:stage_success}
    \vspace{-4pt}
\end{figure}

\begin{figure*}[!t]
    \centering
    \includegraphics[width=\textwidth]{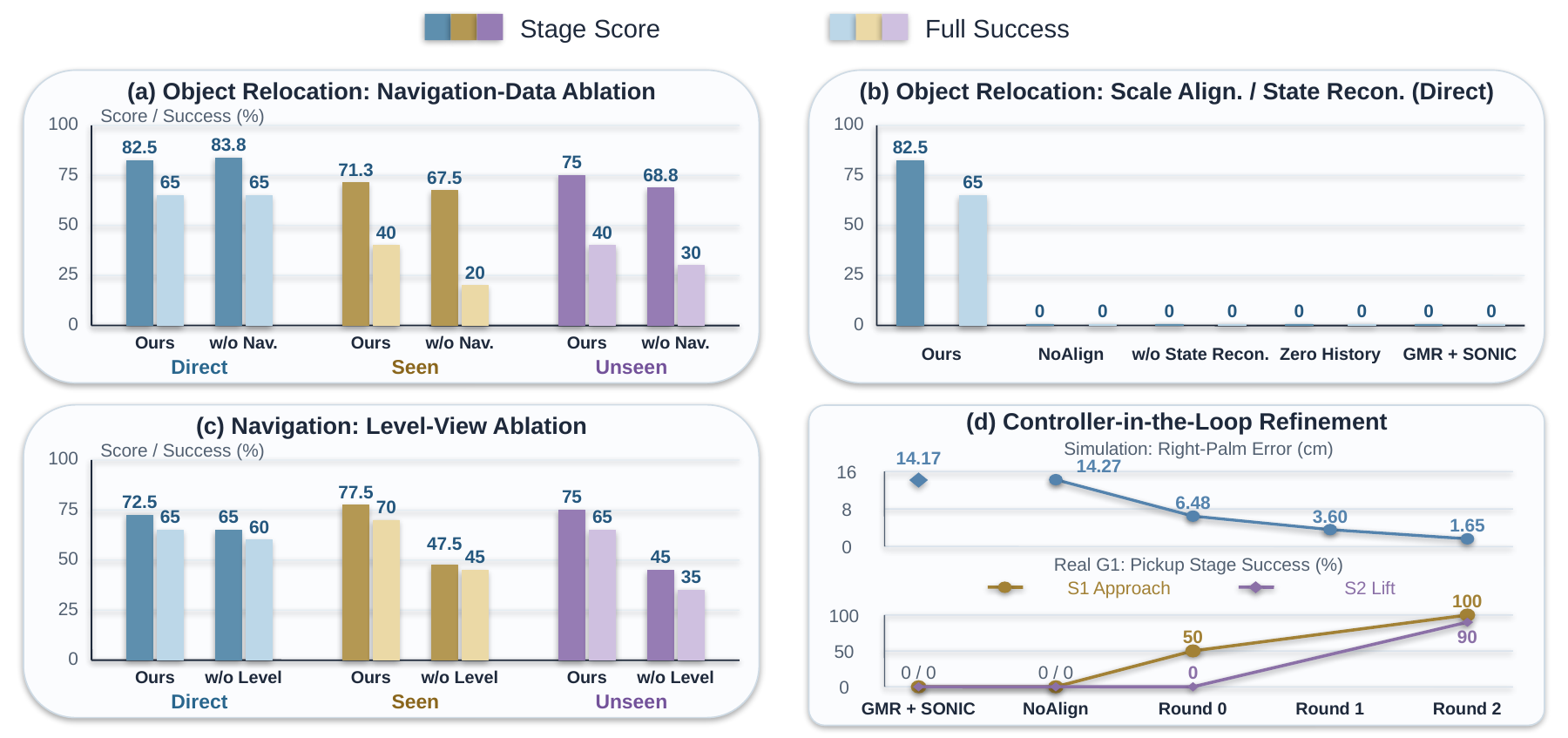}
    \par\vspace{-3pt}
    \caption{\centering Ablations of (a) navigation data, (b) alignment/state reconstruction,
    and (c) level view. (d) Refinement: simulated right-palm error
    (macro mean and 95\% bootstrap CI) and physical S1/S2 success
    (20 Direct trials each for NoAlign and Rounds~0/2). GMR--SONIC fails at S1. Round~1 has simulation results only.}
    \label{fig:data_ablation}
    \vspace{-8pt}
\end{figure*}

Table~\ref{tab:main_results} reports main and ablation results. Fig.~\ref{fig:stage_success} summarizes the full-method navigation and foot interaction subtasks, while Fig.~\ref{fig:data_ablation}(a) compares object relocation with its navigation-data ablation.

\begin{table*}[!t]
\caption{Main and ablation results on G1 (20 trials per row).
Score: mean stage completion; Full Suc.: final-stage success.
Round~0 evaluates pickup only; dashes indicate unreported entries.}
\label{tab:main_results}
\centering
\scriptsize
\setlength{\tabcolsep}{5.2pt}
\begin{tabular}{lllccccccc}
\hline
Task / Subtask & Endpoint & Variant & Trials & Score & Full Suc. & S1 & S2 & S3 & S4 \\
\hline
\textbf{Object Relocation} & \textbf{Direct} & \textbf{Aligned (ours)} & \textbf{20} & \textbf{82.5} & \textbf{65} & \textbf{100} & \textbf{90} & \textbf{75} & \textbf{65} \\
 & & w/o Nav. data & 20 & 83.8 & 65 & 100 & 95 & 75 & 65 \\
 & & NoAlign & 20 & 0.0 & 0 & 0 & 0 & 0 & 0 \\
 & & Kinematic (Round 0) & 20 & -- & -- & 50 & 0 & -- & -- \\
 & & w/o State Recon. & 20 & 0.0 & 0 & 0 & 0 & 0 & 0 \\
 & & Zero History & 20 & 0.0 & 0 & 0 & 0 & 0 & 0 \\
\cline{2-10}
 & \textbf{Multi-seen} & \textbf{Aligned (ours)} & \textbf{20} & \textbf{71.3} & \textbf{40} & \textbf{100} & \textbf{90} & \textbf{55} & \textbf{40} \\
 & & w/o Nav. data & 20 & 67.5 & 20 & 100 & 90 & 60 & 20 \\
\cline{2-10}
 & \textbf{Multi-unseen} & \textbf{Aligned (ours)} & \textbf{20} & \textbf{75.0} & \textbf{40} & \textbf{100} & \textbf{90} & \textbf{70} & \textbf{40} \\
 & & w/o Nav. data & 20 & 68.8 & 30 & 100 & 85 & 60 & 30 \\
\hline
\textbf{Navigation} & \textbf{Direct} & \textbf{Dual-view (ours)} & \textbf{20} & \textbf{72.5} & \textbf{65} & \textbf{80} & \textbf{65} & \textbf{--} & \textbf{--} \\
 & & Downward-only & 20 & 65.0 & 60 & 70 & 60 & -- & -- \\
\cline{2-10}
 & \textbf{Multi-seen} & \textbf{Dual-view (ours)} & \textbf{20} & \textbf{77.5} & \textbf{70} & \textbf{85} & \textbf{70} & \textbf{--} & \textbf{--} \\
 & & Downward-only & 20 & 47.5 & 45 & 50 & 45 & -- & -- \\
\cline{2-10}
 & \textbf{Multi-unseen} & \textbf{Dual-view (ours)} & \textbf{20} & \textbf{75.0} & \textbf{65} & \textbf{85} & \textbf{65} & \textbf{--} & \textbf{--} \\
 & & Downward-only & 20 & 45.0 & 35 & 55 & 35 & -- & -- \\
\hline
Foot Interaction & Multi-seen & Ours & 20 & 80.0 & 60 & 100 & 80 & 60 & -- \\
 & Multi-unseen & Ours & 20 & 78.3 & 60 & 100 & 75 & 60 & -- \\
\hline
\end{tabular}
\vspace{-4pt}
\end{table*}

Object relocation attains 65\% full success on Direct and 40\% on both Multi-seen and Multi-unseen. Despite 90\% pickup success in both multi-position protocols, subsequent turning errors during transport caused the robot to leave the test area or fail to reach the table for placement. Navigation achieves 65--70\% success across Direct, Multi-seen, and Multi-unseen settings. The foot interaction subtask attains 60\% full success for both Multi-seen and Multi-unseen endpoints. Overall, the observed performance is similar on seen and unseen endpoints. These results show that the constructed human supervision supports learning long-range loco-manipulation and navigation to held-out goal positions, as well as leg motions for independently evaluated foot interaction.

\subsection{Human Data-Collection Efficiency}

Table~\ref{tab:collection_efficiency} compares the on-site time required to acquire one object relocation demonstration. Human collection reduces the recording-and-reset cycle from 130~s to 25~s, increasing throughput by $5.2\times$. Times average 10 demonstrations per collection mode, all valid without re-recording, and cover recording and scene reset. One-time equipment setup takes 30 minutes for teleoperation and 5 minutes for human collection, excluded from these per-demonstration times.

\begin{table}[t]
\caption{On-site object relocation data-acquisition time per demonstration.}
\label{tab:collection_efficiency}
\centering
\scriptsize
\setlength{\tabcolsep}{4.2pt}
\begin{tabular}{lrrrr}
\hline
Data & Demo & Reset & Total & Rate \\
\hline
Teleop & 70s & 60s & 130s & $1.0\times$ \\
Human ego & 15s & 10s & 25s & $5.2\times$ \\
\hline
\end{tabular}
\vspace{-8pt}
\end{table}

\subsection{Ablation Study}

We evaluate component contributions within the SONIC stack and compare complete data-construction pipelines using GMR--SONIC. Fig.~\ref{fig:data_ablation}(a--c) compares policy ablations under the protocols in Table~\ref{tab:main_results}. Compared variants are constructed from the same selected training source demonstrations, except that the navigation-data ablation removes the additional navigation demonstrations. All four comparison methods in Fig.~\ref{fig:data_ablation}(b) fail at S1 in every Direct trial and are therefore not evaluated on the multi-position settings.

\subsubsection{Navigation Data}

This ablation removes the 100 additional basket-carrying navigation demonstrations while retaining all 200 full trajectories and the remaining pipeline. The full method and \emph{w/o Navigation Data} perform comparably on Direct, both achieving 65\% full success. With navigation data, full success is 40\% on both multi-position settings, compared with 20\% and 30\% without it on Multi-seen and Multi-unseen, respectively. Successful placement (S4) requires navigating accurately to the placement location; with added navigation demonstrations, we observe higher success rates on the multi-position settings and comparable performance on Direct.

\subsubsection{Downward-View and Level-View Perception}

To measure the value of distant context, \emph{Downward-only} removes the level-view stream during both fine-tuning and deployment, whereas \emph{Dual-view} uses the paired streams. Both variants use identical motion and state data and achieve comparable full success on Direct. Removing the level view reduces full success by 25 and 30 percentage points on Multi-seen and Multi-unseen, respectively. The marked S1 decline in both multi-position protocols supports the role of distant visual context in reaching spatially varied targets (Fig.~\ref{fig:data_ablation}(c)).

\subsubsection{Causal State Reconstruction}

The \emph{w/o State Recon.} object relocation ablation retains the images and scale-aligned human references but zeros the SONIC decoder's history $\mathcal{H}^{G}_{t}$ during training-data construction. The VLA receives the resulting 43-D robot states, not zeroed state inputs. Token labels are recomputed offline using reference-based orientation anchoring instead of the rollout robot-base orientation. \emph{Zero History} uses the same history intervention but retains the full method's rollout token labels. Both use live VLA states and SONIC state history at deployment, as in the full method, and pass S1 in 0/20 Direct trials versus 20/20 for the full method (Fig.~\ref{fig:data_ablation}(b)). The first comparison supports the complete reconstruction pipeline over the tested alternative; the second supports history-conditioned state construction with fixed token supervision.

\subsubsection{Scale Alignment}
\label{sec:scale_alignment_exp}

NoAlign bypasses all alignment, including the left-arm prior and spine constraints. Operators observed a downward hand offset exceeding 10 cm, causing contact with the basket or table rather than entry into the handle gap.
Round~0 uses \emph{kinematic alignment only}: it retains the same arm prior and spine constraints as the full method but omits refinement. All endpoint simulation experiments start from the same complete set of pre-screened object relocation source trajectories, including those excluded from training by final alignment acceptance. A shared target-derived mask selects active frames based on right-palm motion or displacement from its initial position. For NoAlign and Rounds~0--2, we average framewise errors within each trajectory and then equally across trajectories; 95\% percentile bootstrap intervals use 10,000 trajectory-level resamples. With lag compensation, simulated macro error across common active right-palm frames falls from $14.27$ cm (NoAlign) to $6.48$, $3.60$, and $1.65$ cm (Rounds~0--2): a $74.5\%$ reduction beyond kinematic alignment alone (Fig.~\ref{fig:data_ablation}(d)). All rounds are evaluated without early acceptance.

In 20 Direct trials per setting, S1/S2 success is 0\%/0\% for NoAlign, 50\%/0\% for a policy trained on Round~0 supervision, and 100\%/90\% for the full method (Round~2); Round~0 trials end after S2 (Table~\ref{tab:main_results}). Refinement thus improves physical approach and pickup alongside simulated endpoint accuracy.

We also compare a GMR~\cite{araujo2025retargeting}--SONIC pipeline: GMR uses matched palm targets and G1 geometry, then feeds joint references through SONIC without post-decoder overrides or refinement. Simulated rollouts complete; the paired comparison includes only trajectories with complete coverage of the fixed active-frame mask after temporal alignment for all compared methods. Right-palm macro error is $14.17$ cm for GMR--SONIC, $6.49$ cm for kinematic alignment only, and $1.65$ cm for EgoAlign. All physical GMR--SONIC object relocation trials fail at S1 (Fig.~\ref{fig:data_ablation}(b)).

Without lag compensation, we additionally compare target and realized endpoints at the same control tick over the full shared source set, with the active-frame mask unchanged. Macro mean errors are $14.43$, $7.14$, $4.76$, and $3.59$ cm for NoAlign and Rounds~0--2, respectively, and $14.09$ cm for GMR--SONIC. Thus, the refinement gains and EgoAlign's lower error than GMR--SONIC also hold under same-time evaluation.

\section{DISCUSSION}
\label{sec:discussion}

After two refinement rounds, the simulated right-palm endpoint error averages $1.65$ cm with lag compensation. The residual error highlights the need for further endpoint refinement for precision-sensitive hand operations. The evaluated foot operation is pedal pressing, whose comparatively large contact region accommodates greater positioning error. The physical results support simplified replay as a practical approximation for constructing training supervision in these light-load, contact-tolerant tasks, rather than an exact reconstruction of object-contact dynamics.

\section{CONCLUSION}

EgoAlign bridges the human--humanoid gap through embodiment-aware training-data construction. Motion adaptation accounts for differences in body scale and controller response, while causal state reconstruction pairs the adapted actions with their corresponding robot states. Human-only task training followed by zero-shot physical deployment demonstrates that the resulting supervision supports long-range loco-manipulation through a continuous whole-body interface. The significance lies in connecting the accessibility of egocentric human demonstrations with their training value for a target humanoid: human experience becomes usable supervision through adaptation to the robot's motion and state interfaces.


\input{acknowledgments}

\bibliographystyle{ieeetr_etal}
\bibliography{reference}

\end{document}

%% file: authors.tex
\author{Yiming Jiang$^{1,4,*}$, Jin Chen$^{2,4}$,
Chongyang Xu$^{3,4}$, Yilun Chen$^{4,\dagger}$,\\
Aimin Hao$^{1,\ddagger}$, Yisheng He$^{4,\dagger,\ddagger}$%
\thanks{$^{1}$Beihang University. $^{2}$Shanghai Innovation Institute.
$^{3}$Sichuan University. $^{4}$Alibaba Group.}%
\thanks{ 
$^{\dagger}$Co-project leaders.
$^{\ddagger}$Co-corresponding authors.}%
\thanks{Contact: Yiming Jiang (\texttt{jiangyimingjym@buaa.edu.cn});
Yisheng He (\texttt{ethanheysh@gmail.com}).}%
\thanks{$^{*}$Work done during an internship at Alibaba Token Hub (ATH), Alibaba Group.}%
}

%% file: acknowledgments.tex
\section*{ACKNOWLEDGMENTS}
This work was supported by Alibaba Research Intern Program.

%% file: reference.bib
@article{loper2015smpl,
  author  = {Matthew Loper and Naureen Mahmood and Javier Romero and Gerard Pons-Moll and Michael J. Black},
  title   = {{SMPL}: A Skinned Multi-Person Linear Model},
  journal = {ACM Transactions on Graphics},
  volume  = {34},
  number  = {6},
  pages   = {248:1--248:16},
  year    = {2015},
  doi     = {10.1145/2816795.2818013}
}

@inproceedings{todorov2012mujoco,
  author    = {Emanuel Todorov and Tom Erez and Yuval Tassa},
  title     = {{MuJoCo}: A Physics Engine for Model-Based Control},
  booktitle = {IROS},
  pages     = {5026--5033},
  year      = {2012},
  doi       = {10.1109/IROS.2012.6386109}
}

@inproceedings{kim2024openvla,
  author    = {Kim, Moo Jin and Pertsch, Karl and Karamcheti, Siddharth and Xiao, Ted and Balakrishna, Ashwin and Nair, Suraj and Rafailov, Rafael and Foster, Ethan P and Sanketi, Pannag R and Vuong, Quan and Kollar, Thomas and Burchfiel, Benjamin and Tedrake, Russ and Sadigh, Dorsa and Levine, Sergey and Liang, Percy and Finn, Chelsea},
  title     = {{OpenVLA}: An Open-Source Vision-Language-Action Model},
  booktitle = {CoRL},
  pages     = {2679--2713},
  year      = {2025},
  volume    = {270},
  url       = {https://proceedings.mlr.press/v270/kim25c.html}
}

@article{starvla2026,
  author  = {{StarVLA Community}},
  title   = {{StarVLA}: A Lego-like Codebase for Vision-Language-Action Model Developing},
  journal = {arXiv preprint arXiv:2604.05014},
  year    = {2026},
  url     = {https://arxiv.org/abs/2604.05014}
}

@inproceedings{physicalintelligence2025pi05,
  author    = {Black, Kevin and Brown, Noah and Darpinian, James and Dhabalia, Karan and Driess, Danny and Esmail, Adnan and Equi, Michael Robert and Finn, Chelsea and Fusai, Niccolo and Galliker, Manuel Y. and Ghosh, Dibya and Groom, Lachy and Hausman, Karol and Ichter, Brian and Jakubczak, Szymon and Jones, Tim and Ke, Liyiming and LeBlanc, Devin and Levine, Sergey and Li-Bell, Adrian and Mothukuri, Mohith and Nair, Suraj and Pertsch, Karl and Ren, Allen Z. and Shi, Lucy Xiaoyang and Smith, Laura and Springenberg, Jost Tobias and Stachowicz, Kyle and Tanner, James and Vuong, Quan and Walke, Homer and Walling, Anna and Wang, Haohuan and Yu, Lili and Zhilinsky, Ury},
  title     = {{$\pi_{0.5}$}: A Vision-Language-Action Model with Open-World Generalization},
  booktitle = {CoRL},
  pages     = {17--40},
  year      = {2025},
  volume    = {305},
  url       = {https://proceedings.mlr.press/v305/black25a.html}
}

@inproceedings{black2024pi0,
  author    = {Kevin Black and Noah Brown and Danny Driess and Adnan Esmail and Michael Robert Equi and Chelsea Finn and Niccolo Fusai and Lachy Groom and Karol Hausman and Brian Ichter and Szymon Jakubczak and Tim Jones and Liyiming Ke and Sergey Levine and Adrian Li-Bell and Mohith Mothukuri and Suraj Nair and Karl Pertsch and Lucy Xiaoyang Shi and Laura Smith and James Tanner and Quan Vuong and Anna Walling and Haohuan Wang and Ury Zhilinsky},
  title     = {{$\pi_0$}: A Vision-Language-Action Flow Model for General Robot Control},
  booktitle = {RSS},
  year      = {2025},
  doi       = {10.15607/RSS.2025.XXI.010},
  url       = {https://www.roboticsproceedings.org/rss21/p010.html}
}

@article{physicalintelligence2026pi07,
  author  = {{Physical Intelligence} and Bo Ai and Ali Amin and Raichelle Aniceto and Ashwin Balakrishna and Greg Balke and others},
  title   = {{$\pi_{0.7}$}: A Steerable Generalist Robotic Foundation Model with Emergent Capabilities},
  journal = {arXiv preprint arXiv:2604.15483},
  year    = {2026},
  url     = {https://arxiv.org/abs/2604.15483}
}

@article{zhang2026joyaira,
  author  = {Tianle Zhang and Zhihao Yuan and Dafeng Chi and Peidong Liu and Dongwei Li and Kejun Hu and others},
  title   = {{JoyAI-RA 0.1}: A Foundation Model for Robotic Autonomy},
  journal = {arXiv preprint arXiv:2604.20100},
  year    = {2026},
  url     = {https://arxiv.org/abs/2604.20100}
}

@inproceedings{wei2026psi0,
  author    = {Songlin Wei and Hongyi Jing and Boqian Li and Zhenyu Zhao and Jiageng Mao and Zhenhao Ni and Sicheng He and Sheng Zang and Xiawei Liu and Kaidi Kang and Jie Liu and Weiduo Yuan and Marco Pavone and Di Huang and Yue Wang},
  title     = {{$\Psi_0$}: An Open Foundation Model Towards Universal Humanoid Loco-Manipulation},
  booktitle = {RSS},
  year      = {2026},
  doi       = {10.15607/RSS.2026.XXII.021},
  url       = {https://www.roboticsproceedings.org/rss22/p021.html}
}

@inproceedings{chi2024umi,
  author    = {Cheng Chi and Zhenjia Xu and Chuer Pan and Eric Cousineau and Benjamin Burchfiel and Siyuan Feng and Russ Tedrake and Shuran Song},
  title     = {Universal Manipulation Interface: In-the-Wild Robot Teaching Without In-the-Wild Robots},
  booktitle = {RSS},
  year      = {2024},
  doi       = {10.15607/RSS.2024.XX.045},
  url       = {https://www.roboticsproceedings.org/rss20/p045.html}
}

@inproceedings{zhaxizhuoma2025fastumi,
  author    = {Zhaxizhuoma and Kehui Liu and Chuyue Guan and Zhongjie Jia and Ziniu Wu and Xin Liu and Tianyu Wang and Shuai Liang and Pengan Chen and Pingrui Zhang and Haoming Song and Delin Qu and Dong Wang and Zhigang Wang and Nieqing Cao and Yan Ding and Bin Zhao and Xuelong Li},
  title     = {{FastUMI}: A Scalable and Hardware-Independent Universal Manipulation Interface with Dataset},
  booktitle = {CoRL},
  volume    = {305},
  pages     = {3069--3093},
  year      = {2025},
  url       = {https://proceedings.mlr.press/v305/zhaxizhuoma25a.html}
}

@article{li2026egolive,
  author  = {Yihang Li and Xuelong Wei and Jingzhou Luo and Yingjing Xiao and Yibo Bai and Guangyuan Zhou and others},
  title   = {{EgoLive}: A Large-Scale Egocentric Dataset from Real-World Human Tasks},
  journal = {arXiv preprint arXiv:2604.23570},
  year    = {2026},
  url     = {https://arxiv.org/abs/2604.23570}
}

@inproceedings{punamiya2026egoverse,
  author    = {Ryan Punamiya and Simar Kareer and Zeyi Liu and Joshua Citron and Ri-Zhao Qiu and Xiongyi Cai and Alexey Gavryushin and Jiaqi Chen and Davide Liconti and Lawrence Y. Zhu and Patcharapong Aphiwetsa and Baoyu Li and Aniketh Cheluva and Pranav Kuppili and Yangcen Liu and Dhruv Patel and Aidan Gao and Ryan Co and Hye-Young Chung and Renee Zbizika and Jinyun Liu and Xiaomeng Xu and Haoyu Xiong and Geng Chen and Sebastiano Oliani and Wenkai Xuan and Chenyu Yang and Xi Wang and James Fort and Richard Newcombe and Josh Gao and Jason Chong and Garrett Matsuda and Aseem Doriwala and Robert K. Katzschmann and Marc Pollefeys and Xiaolong Wang and Shuran Song and Judy Hoffman and Danfei Xu},
  title     = {{EgoVerse}: An Egocentric Human Dataset for Robot Learning from Around the World},
  booktitle = {RSS},
  year      = {2026},
  doi       = {10.15607/RSS.2026.XXII.092},
  url       = {https://www.roboticsproceedings.org/rss22/p092.html}
}

@article{zheng2026egoscale,
  author  = {Ruijie Zheng and Dantong Niu and Yuqi Xie and Jing Wang and Mengda Xu and Yunfan Jiang and Fernando Casta{\~n}eda and Fengyuan Hu and You Liang Tan and Letian Fu and Trevor Darrell and Furong Huang and Yuke Zhu and Danfei Xu and Linxi Fan},
  title   = {{EgoScale}: Scaling Dexterous Manipulation with Diverse Egocentric Human Data},
  journal = {arXiv preprint arXiv:2602.16710},
  year    = {2026},
  url     = {https://arxiv.org/abs/2602.16710}
}

@inproceedings{kareer2024egomimic,
  author    = {Simar Kareer and Dhruv Patel and Ryan Punamiya and Pranay Mathur and Shuo Cheng and Chen Wang and Judy Hoffman and Danfei Xu},
  title     = {{EgoMimic}: Scaling Imitation Learning via Egocentric Video},
  booktitle = {ICRA},
  pages     = {13226--13233},
  year      = {2025},
  doi       = {10.1109/ICRA55743.2025.11127989},
  url       = {https://ieeexplore.ieee.org/document/11127989}
}

@article{wang2026humanego,
  author  = {Zhi Wang and Botao He and Kelin Yu and Seungjae Lee and Ruohan Gao and Furong Huang and Yiannis Aloimonos},
  title   = {{HumanEgo}: Zero-Shot Robot Learning from Minutes of Human Egocentric Videos},
  journal = {arXiv preprint arXiv:2605.24934},
  year    = {2026},
  url     = {https://arxiv.org/abs/2605.24934}
}

@article{wang2026ego2robot,
  author  = {Ye Wang and Pei Lin and Xiong-Hui Chen and Haoqi Yuan and Zhixuan Liang and Yiyang Huang and Anzhe Chen and Zixing Lei and Jie Zhang and Tao Zhang and Haoyang Li and Tong Zhang and Chenxi Xiao and Ziyuan Jiao and Qin Jin},
  title   = {{Ego2Robot}: Scalable Robot Data Synthesis from Egocentric Human Data},
  journal = {arXiv preprint arXiv:2608.02580},
  year    = {2026},
  url     = {https://arxiv.org/abs/2608.02580}
}

@inproceedings{he2024omnih2o,
  author    = {He, Tairan and Luo, Zhengyi and He, Xialin and Xiao, Wenli and Zhang, Chong and Zhang, Weinan and Kitani, Kris M. and Liu, Changliu and Shi, Guanya},
  title     = {{OmniH2O}: Universal and Dexterous Human-to-Humanoid Whole-Body Teleoperation and Learning},
  booktitle = {CoRL},
  pages     = {1516--1540},
  year      = {2025},
  volume    = {270},
  url       = {https://proceedings.mlr.press/v270/he25b.html}
}

@inproceedings{ben2025homie,
  author    = {Qingwei Ben and Feiyu Jia and Jia Zeng and Junting Dong and Dahua Lin and Jiangmiao Pang},
  title     = {{HOMIE}: Humanoid Loco-Manipulation with Isomorphic Exoskeleton Cockpit},
  booktitle = {RSS},
  year      = {2025},
  doi       = {10.15607/RSS.2025.XXI.070},
  url       = {https://www.roboticsproceedings.org/rss21/p070.html}
}

@article{nai2026humi,
  author  = {Ruiqian Nai and Boyuan Zheng and Junming Zhao and Haodong Zhu and Sicong Dai and Zunhao Chen and Yihang Hu and Yingdong Hu and Tong Zhang and Chuan Wen and Yang Gao},
  title   = {Humanoid Manipulation Interface: Humanoid Whole-Body Manipulation from Robot-Free Demonstrations},
  journal = {arXiv preprint arXiv:2602.06643},
  year    = {2026},
  url     = {https://arxiv.org/abs/2602.06643}
}

@article{wang2026bifrostumi,
  author  = {Hongwu Wang and Chenhao Yu and Youhao Hu and Jiachen Zhang and Yuanyuan Li and Shaqi Luo},
  title   = {{BifrostUMI}: Bridging Robot-Free Demonstrations and Humanoid Whole-Body Manipulation},
  journal = {arXiv preprint arXiv:2605.03452},
  year    = {2026},
  url     = {https://arxiv.org/abs/2605.03452}
}

@inproceedings{shi2026egohumanoid,
  author    = {Modi Shi and Shijia Peng and Jin Chen and Haoran Jiang and Tianyu Li and Ping Luo and Di Huang and Hongyang Li and Li Chen},
  title     = {Unlocking In-the-Wild Loco-Manipulation with Robot-Free Egocentric Demonstration},
  booktitle = {RSS},
  year      = {2026},
  doi       = {10.15607/RSS.2026.XXII.204},
  url       = {https://www.roboticsproceedings.org/rss22/p204.html}
}

@article{luo2025sonic,
  author    = {Zhengyi Luo and Ye Yuan and Tingwu Wang and Chenran Li and Fernando Casta{\~n}eda and Sirui Chen and Zi-Ang Cao and Jiefeng Li and David Minor and Qingwei Ben and Jinhyung Park and David Sami and Zi Wang and Xingye Da and Runyu Ding and Cyrus Hogg and Lina Song and Edy Lim and Eugene Jeong and Tairan He and Haoru Xue and Wenli Xiao and Simon Yuen and Jan Kautz and Yan Chang and Umar Iqbal and Linxi ``Jim'' Fan and Yuke Zhu},
  title     = {{SONIC}: Supersizing Motion Tracking for Natural Humanoid Whole-Body Control},
  journal   = {Science Robotics},
  volume    = {11},
  number    = {117},
  note      = {Art. no. eaed4592},
  year      = {2026},
  doi       = {10.1126/scirobotics.aed4592},
  url       = {https://www.science.org/doi/10.1126/scirobotics.aed4592}
}

@article{li2026tango,
  author  = {Anqi Li and Yuxin Chen and Zhaobo Li and Zhuo Cao and Junli Ren and Masayoshi Tomizuka and Dhruv Shah},
  title   = {{TANGO}: Humanoid Navigation in Cluttered Environments with a Whole-Body Vision-Language-Action Model},
  journal = {arXiv preprint arXiv:2609.09158},
  year    = {2026},
  url     = {https://arxiv.org/abs/2609.09158}
}

@article{wang2026vlk,
  author  = {Yen-Jen Wang and Jiaman Li and Sirui Chen and Takara E. Truong and Pei Xu and Pieter Abbeel and Rocky Duan and Koushil Sreenath and Angjoo Kanazawa and Carmelo Sferrazza and Guanya Shi and Karen Liu},
  title   = {{VLK}: Learning Humanoid Loco-Manipulation from Synthetic Interactions in Reconstructed Scenes},
  journal = {arXiv preprint arXiv:2606.30645},
  year    = {2026},
  url     = {https://arxiv.org/abs/2606.30645}
}

@article{deng2026prosgnerf,
  title     = {{ProSGNeRF}: Progressive Dynamic Neural Scene Graph with Frequency Modulated Foundation Model in Urban Scenes},
  author    = {Deng, Tianchen and Wang, Yanbo and Liu, Yejia and Su, Chenpeng and Wang, Jingchuan and Wang, Hesheng and Wang, Danwei and Lo, Shao-Yuan and Chen, Weidong},
  journal   = {International Journal of Computer Vision},
  volume    = {134},
  number    = {8},
  note      = {Art. no. 365},
  year      = {2026},
  doi       = {10.1007/s11263-026-02962-5},
  url       = {https://link.springer.com/article/10.1007/s11263-026-02962-5}
}

@article{nie2026towards,
  title   = {Towards the Vision-Sound-Language-Action Paradigm: The {HEAR} Framework for Sound-Centric Manipulation},
  author  = {Nie, Chang and Deng, Tianchen and Wang, Guangming and Liu, Zhe and Wang, Hesheng},
  journal = {arXiv preprint arXiv:2603.16086},
  year    = {2026},
  url     = {https://arxiv.org/abs/2603.16086}
}

@inproceedings{araujo2025retargeting,
  author    = {Ara{\'u}jo, Jo{\~a}o Pedro and Ze, Yanjie and Xu, Pei and Wu, Jiajun and Liu, C. Karen},
  title     = {Retargeting Matters: General Motion Retargeting for Humanoid Motion Tracking},
  booktitle = {ICRA},
  year      = {2026},
  url       = {https://jiajunwu.com/papers/gmr_icra.pdf}
}

@article{beyer2024paligemma,
  title   = {{PaliGemma}: A Versatile {3B VLM} for Transfer},
  author  = {Lucas Beyer and Andreas Steiner and Andr{\'e} Susano Pinto and Alexander Kolesnikov and Xiao Wang and Daniel Salz and Maxim Neumann and Ibrahim Alabdulmohsin and Michael Tschannen and Emanuele Bugliarello and Thomas Unterthiner and Daniel Keysers and Skanda Koppula and Fangyu Liu and Adam Grycner and Alexey Gritsenko and Neil Houlsby and Manoj Kumar and Keran Rong and Julian Eisenschlos and Rishabh Kabra and Matthias Bauer and Matko Bo{\v{s}}njak and Xi Chen and Matthias Minderer and Paul Voigtlaender and Ioana Bica and Ivana Balazevic and Joan Puigcerver and Pinelopi Papalampidi and Olivier Henaff and Xi Xiong and Radu Soricut and Jeremiah Harmsen and Xiaohua Zhai},
  journal = {arXiv preprint arXiv:2407.07726},
  year    = {2024},
  url     = {https://arxiv.org/abs/2407.07726}
}

@article{deng2025best3dscenerepresentation,
  title   = {What Is the Best {3D} Scene Representation for Robotics? {From} Geometric to Foundation Models},
  author  = {Deng, Tianchen and Pan, Yue and Yuan, Shenghai and Li, Dong and Wang, Chen and Li, Mingrui and Chen, Long and Xie, Lihua and Wang, Danwei and Wang, Jingchuan and Civera, Javier and Wang, Hesheng and Chen, Weidong},
  journal = {arXiv preprint arXiv:2512.03422},
  year    = {2025},
  url     = {https://arxiv.org/abs/2512.03422}
}
